\documentclass{article}

\usepackage{microtype}
\usepackage{graphicx}
\usepackage{subfigure}
\usepackage{booktabs} 

\usepackage{hyperref}

\usepackage[accepted]{icml2024}

\usepackage{amsmath}
\usepackage{amssymb}
\usepackage{mathtools}
\usepackage{amsthm}
\usepackage{bbding}
\usepackage{multirow} 
\usepackage[capitalize,noabbrev]{cleveref}

\theoremstyle{plain}

\theoremstyle{definition}

\theoremstyle{remark}

\usepackage[textsize=tiny]{todonotes}

\icmltitlerunning{Adaptive Feature Selection for No-Reference Image Quality Assessment by Mitigating Semantic Noise Sensitivity}

\begin{document}

\twocolumn[
\icmltitle{Adaptive Feature Selection for No-Reference Image Quality Assessment by Mitigating Semantic Noise Sensitivity}



\icmlsetsymbol{equal}{*}

\begin{icmlauthorlist}
\icmlauthor{Xudong Li}{equal,mac}
\icmlauthor{Timin Gao}{equal,mac}
\icmlauthor{Runze Hu}{equal,hrz}
\icmlauthor{Yan Zhang}{mac}
\icmlauthor{Shengchuan Zhang}{mac}
\icmlauthor{Xiawu Zheng}{mac}
\icmlauthor{Jingyuan Zheng}{med}
\icmlauthor{Yunhang Shen}{youtu}
\icmlauthor{Ke Li}{youtu}
\icmlauthor{Yutao Liu}{oca}
\icmlauthor{Pingyang Dai}{mac}
\icmlauthor{Rongrong Ji}{mac}
\end{icmlauthorlist}

\icmlaffiliation{mac}{Key Laboratory of Multimedia Trusted Perception and Efficient Computing, Ministry of Education of China, Xiamen University, 361005, P.R. China.
}
\icmlaffiliation{med}{School of Medicine, Xiamen University}
\icmlaffiliation{hrz}{School of Information and Electronics, Beijing Institute of Technology}
\icmlaffiliation{youtu}{Tencent Youtu Lab}
\icmlaffiliation{oca}{School of Computer Science and Technology, Ocean University of China}

\icmlcorrespondingauthor{Yan Zhang}{bzhy986@gmail.com}

\icmlkeywords{Machine Learning, ICML}

\vskip 0.3in
]



\printAffiliationsAndNotice{\icmlEqualContribution} 

\begin{abstract}
The current state-of-the-art No-Reference Image Quality Assessment (NR-IQA) methods typically rely on feature extraction from upstream semantic backbone networks, assuming that all extracted features are relevant. However, we make a key observation that not all features are beneficial, and some may even be harmful, necessitating careful selection. Empirically, we find that many image pairs with small feature spatial distances can have vastly different quality scores, indicating that the extracted features may contain quality-irrelevant noise. To address this issue, we propose a Quality-Aware Feature Matching IQA Metric~(QFM-IQM) that employs an adversarial perspective to remove harmful semantic noise features from the upstream task. Specifically, QFM-IQM enhances the semantic noise distinguish capabilities by matching image pairs with similar quality scores but varying semantic features as adversarial semantic noise and adaptively adjusting the upstream task’s features by reducing sensitivity to adversarial noise perturbation. Furthermore, we utilize a distillation framework to expand the dataset and improve the model's generalization ability. Extensive experiments conducted on eight standard IQA datasets have demonstrated the effectiveness of our proposed QFM-IQM.
\end{abstract}    
\begin{figure}[t!]
  \centering
    \includegraphics[width=0.47\textwidth]{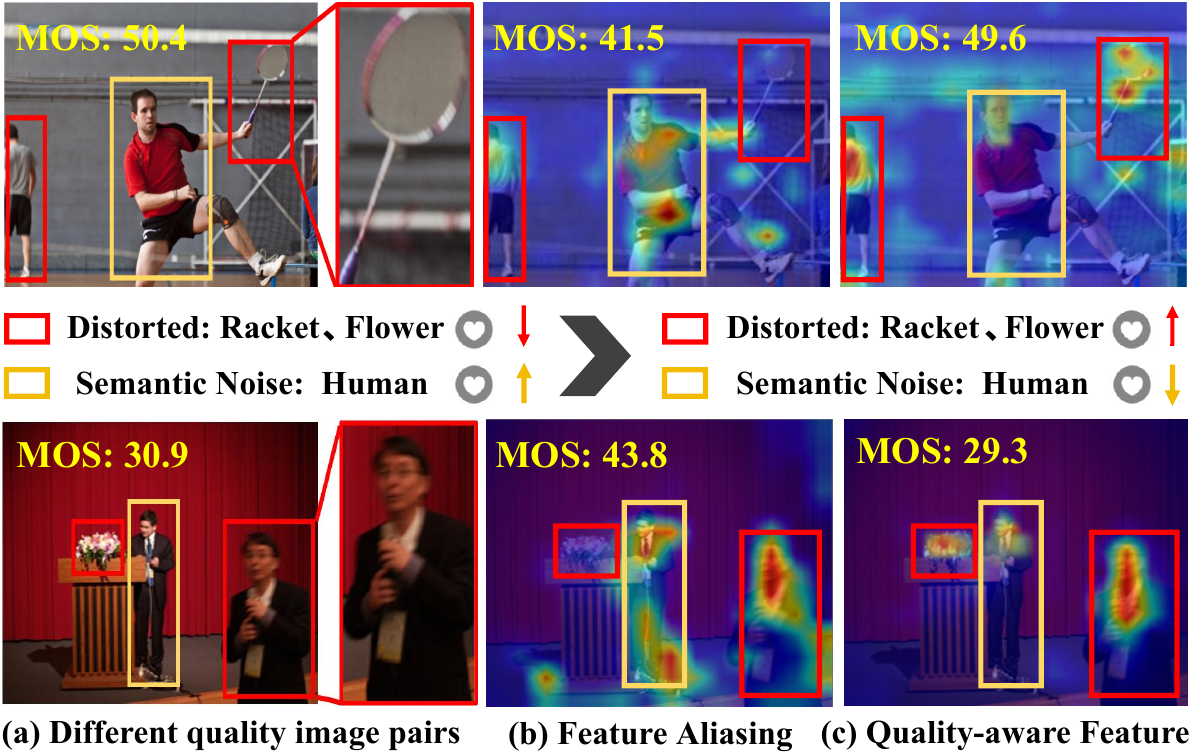}
  \caption{\textbf{Images in the first column:} the sample images in the KonIQ dataset. \textbf{Images in the medium column:} DEIQT~\cite{DEIQT} suffers from feature aliasing that focuses excessively on semantic-level noise that is less relevant to quality perception~(e.g., the human in the yellow box). \textbf{Images in the last column:} Our QFM-IQAM is not sensitive to semantic noise and can clearly distinguish between image pairs with similar semantic content but different quality scores. As a result, our model's predictions are more aligned with the \emph{Mean Opinion Scores} (MOS).}
  \label{fig1}
\end{figure}
\section{Introduction}
    \emph{No-Reference Image Quality Assessment} (NR-IQA) is a fundamental research area~\cite{ding2020image,gu2020giqa,zhang2015som,hu2,hu3}, which simulates the human subjective system to estimate distortion of the given image.
    Current state-of-the-art methods~\cite{zhang2023blind,song2023active} generally leverage pre-trained upstream backbone to extract semantic features, and then, finetuning on NR-IQA datasets.
    This pipeline provides an effective and efficient training procedure while reducing the requirement of data volume.
   Since the semantic features obtained from the pre-trained upstream backbone have no direct correlation with the image quality representation~\cite{yang2020ttl}, many researchers focus on the optimization of semantic features in IQA research.
    For instance, 
    \cite{talebi2018nima} proposed a model based on Deep Convolutional Neural Networks pre-trained on the ImageNet~\cite{imagenet} and end-to-end training to optimize semantic features. ~\cite{hypernet} integrated NR-IQA tasks with semantic recognition networks, enabling the network to assess image quality after identifying content. ~\cite{DEIQT} refined the abstract semantic information obtained from the pre-trained model by introducing a transformer decoder.
    
    However, we empirically find that many image pairs with small feature spatial distances have vastly different quality scores (Sec. \ref{challenge}). We attribute this result to feature aliasing, where semantic noise unrelated to quality confuses the quality-aware features.
    Specifically, due to the sensitivity of upstream backbone networks to semantic features~\cite{zhao2023quality,zhang2023blind}, they tend to extract similar features for semantically similar image pairs, despite these pairs having distinctly different quality scores. This may mislead the network to overly focus on semantic details rather than true indicators of image quality, resulting in poor quality prediction. We provide an example in Fig.\ref{fig1}(b) to illustrate this issue. For a pair of images with obvious quality differences, the baseline overly focuses on similar semantic information in the distorted image (e.g., the person in the yellow box), while ignoring some distorted areas in the red box (e.g., the racket's movement and the flower's overexposure), causing the model to be misled into making a close quality predictions.
    These findings motivate us to explore a critical problem: how to distinguish images with similar semantic information but different quality scores.
    
    We believe the key to addressing this challenge is to reduce the model's sensitivity to semantic noise. Inspired by adversarial learning~\cite{goodfellow2014explaining,madry2017towards}, which aims to learn adversarial noise that maximizes prediction loss, and reduces the model's sensitivity to this noise by minimizing the model's loss when facing noisy inputs. 
    In this paper, we treat samples with similar quality scores but significantly different semantic features compared to the input image as adversarial noise. This approach is based on our hypothesis that the pre-trained upstream backbone is highly sensitive to semantics. Therefore, introducing appropriate semantic perturbation can significantly impact its quality predictions, while similar quality scores ensure that the misprediction is mainly influenced by the semantic perturbation rather than the quality perturbation.
           
   To achieve this, we propose a novel approach called Quality-Aware Feature    Matching Image Quality Metric (QFM-IQM), as shown in Fig.~\ref{fig1}(c). 
    Concretely, our QFM-IQM comprises three primary modules. Firstly, the Semantic Noise Feature Matching (SNM) Module is developed to pair each distorted sample with noise samples that have similar quality scores but significantly different semantic features. The Quality Consistency Constraint (QCC) Module uses noise samples as the adversarial feature perturbation to maintain quality information while disrupting semantics and ensures consistent quality predictions for distorted samples before and after semantic changes. 
    The core objective is to enhance the model’s robustness to semantic noise by minimizing the loss of quality prediction when faced with semantic noise perturbation.  
    Lastly, the Distilled Label Expansion (DLE) Module utilizes knowledge distillation to provide pseudo-labels for unlabeled samples, enriching the dataset for QCC's adversarial learning. 
    Our contributions include:
    \vspace{-5pt}
    \begin{itemize}    
        \item We address a common challenge in IQA: how to distinguish between images that have similar semantic information but different quality scores. To solve this problem, we propose a novel model that can effectively capture the subtle differences in quality among similar images and make accurate judgments based on them.
            \vspace{-3pt}
        \item We introduce a novel feature adversarial learning mechanism to isolate the quality-related attributes from the semantic content of an image. This way, the model can focus on the most relevant attributes for image quality and avoid being distracted by irrelevant ones.
            \vspace{-3pt}
        \item Our model employs distillation learning to augment the dataset with authentic images that have varying quality levels. This not only improves the model’s generalization ability but also reduces the need for human annotation efforts, which are time-consuming.
    \end{itemize}

\begin{figure*}[t]
	\centering{\includegraphics[width=1\textwidth]{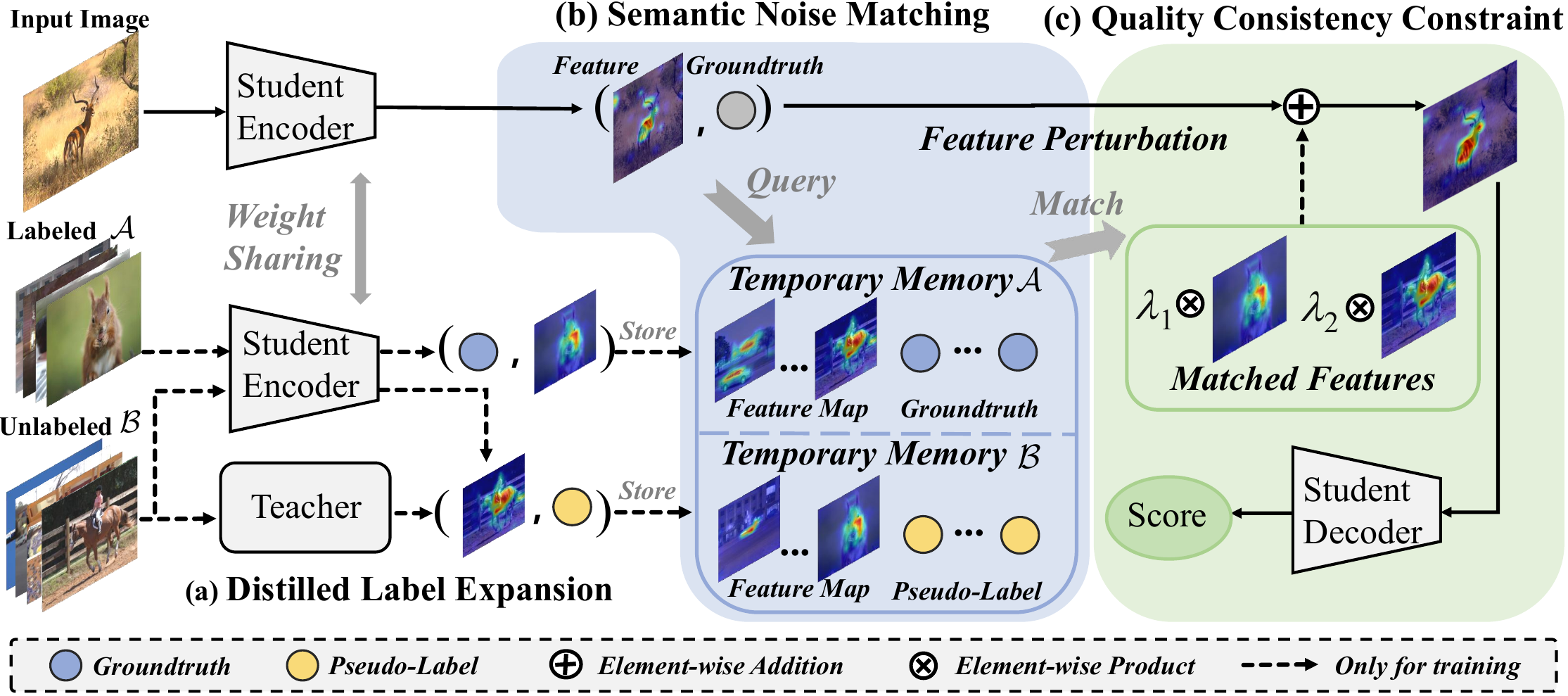}}
    \vspace{-5pt}
	\caption{The overview of our QFM-IQM.We begin by extracting image semantic-related features from the student network. Simultaneously, we indirectly augment our training set by introducing a distillation framework and storing the image features and labels in pairs in temporary memory~(Sec.~\ref{DLE}). We use Semantic Noise Matching to select features in the temporary memory with similar quality scores but different semantic features as noise disturbance~(Sec.~\ref{SNM}) and add them to our original features as decoder input to predict image quality, forcing our model to learn to focus on quality-aware features~(Sec.~\ref{QCC}). During inference, scores are directly predicted using the encoder and decoder, while the above three modules are discarded to avoid any additional computational overhead.
 }
 \label{modelStructure}
\end{figure*}
\vspace{-10pt}
\section{Related Works}
\vspace{-5pt}
Image Quality Assessment (IQA)~\cite{hu1} is broadly classified into Full-Reference (FR)~\cite{cao2022incorporating}, Reduced-Reference (RR)~\cite{tao2009reduced}, and No-Reference (NR) methods. NR-IQA assesses quality independently, offering significant application potential.
\vspace{-5pt}
\subsection{NR-IQA with Vision Transformer}
\vspace{-5pt}
Vision Transformer (ViT)~\cite{ViT} showed promising results on several downstream vision tasks. There were mainly two types of ViT-based NR-IQA methods, including hybrid Transformer ~\cite{TReS} and pure ViT-based Transformer ~\cite{musiq, yang2022maniqa,TIQA}. The hybrid architecture generally combined the CNNs with the Transformer, which was responsible for the local and long-range feature characterization, respectively. For instance, \cite{TReS} proposed to use the multi-scale features extracted from ResNet-50, which were fed to the transformer encoder to produce a non-local representation of the image. 
\cite{musiq} designed a multi-scale ViT-based IQA model to handle the arbitrary size of input images. 
Such a feature was initially designed to describe the image content, and thus the preserved features were mainly related to the higher-level visual abstractions which were not adequate in characterizing the quality-aware features\cite{DEIQT}.
\vspace{-5pt}
\subsection{NR-IQA with Contrastive Learning}
\vspace{-5pt}
A new learning paradigm, contrastive learning, has emerged for learning discriminative representations among samples~\cite{caron2020unsupervised,he2020momentum}. This method aims to learn an embedding space where similar samples are attracted, and dissimilar ones are repelled~\cite{jaiswal2020survey,zhao2023quality}. Recently, contrastive learning in IQA has primarily focused on unsupervised approaches, as seen in key studies like QPT~\cite{zhao2023quality}, which introduced a self-supervised method with quality-aware contrastive learning for BIQA. Re-IQA~\cite{saha2023re} employed a mixture of experts to train separate encoders for content and quality features, and CONTRIQUE~\cite{contrastive} used a deep CNN with a contrastive pairwise objective for IQA. These methods have achieved promising results by leveraging self-supervised training, but the cost of pre-training models was often high. In contrast to contrastive learning aimed at constructing discriminative embedding spaces, we leverage adversarial learning~\cite{goodfellow2014explaining,madry2017towards} to reduce the model's sensitivity to semantic noise. This enhances the model's ability to distinguish between images that are semantically similar but have different quality scores.
\begin{figure}[t]
  \centering
    \includegraphics[width=0.47\textwidth]{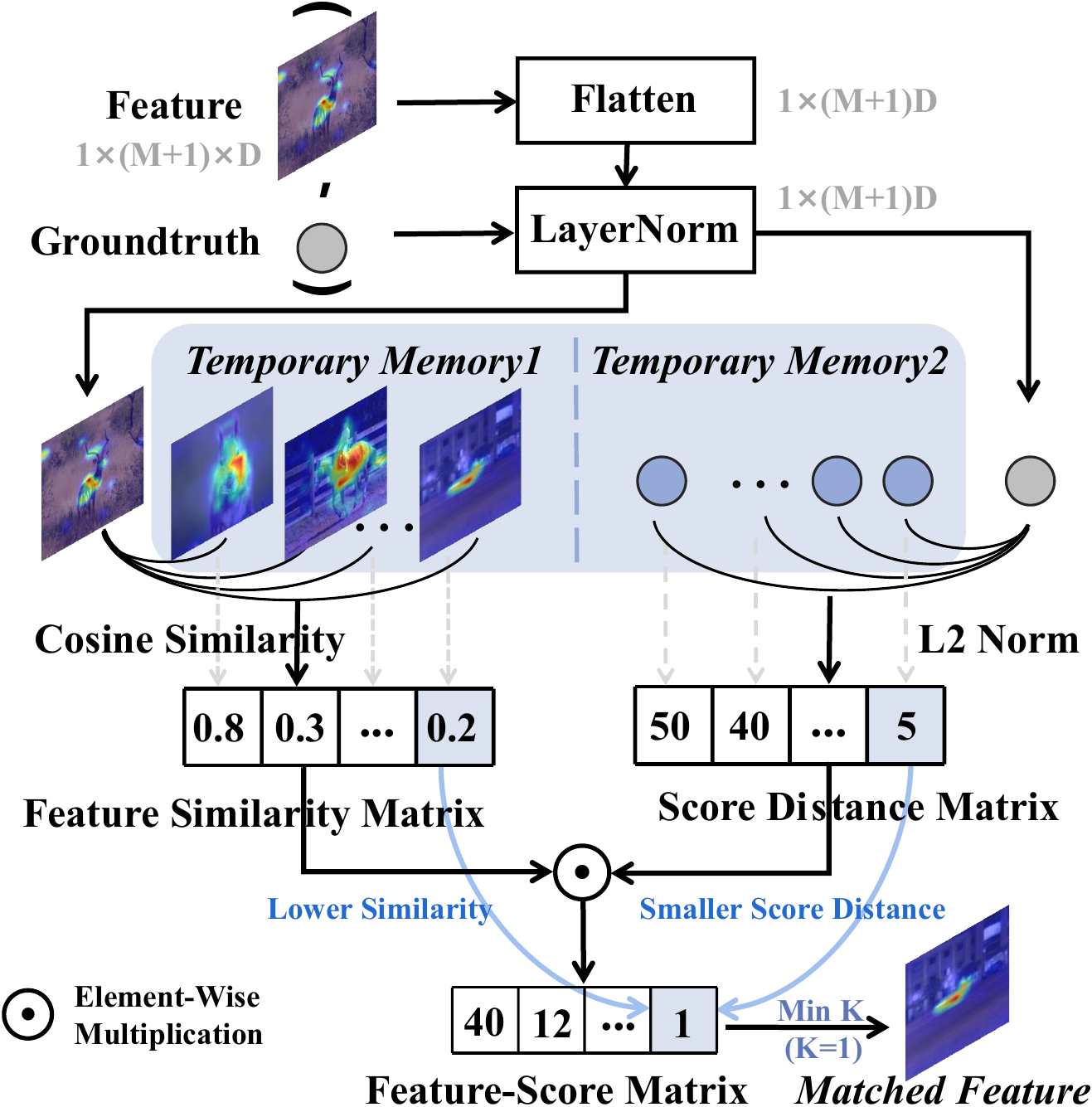}
  \caption{The Semantic Noise Matching~(SNM) pipeline takes in pairs of features and labels as input and returns matching features from the dataset as output. Firstly, we flatten the input features and then normalize both the features and labels using layer norm. We then calculate the Feature Similarity Matrix and the Score Distance Matrix using the cosine similarity function and the $\mathcal L_2$ norm. Next, we obtain the final Feature-Score Matrix by element-wise multiplication of the computed matrices. We select the features that correspond to the index of the minimum K values of the Feature-Score Matrix in our memory as matched features.}
  \vspace{-10pt}
  \label{select}
\end{figure}

\section{Methodology}\label{sec:proposed}
\subsection{Preliminaries}
In the context of No-Reference Image Quality Assessment (NR-IQA), we define some common notations.
We use bold formatting to denote vectors (e.g., $\boldsymbol{x}$, $\boldsymbol{y}$) and matrices (e.g., $\boldsymbol{X}$, $\boldsymbol{Y}$).
Specifically, $\boldsymbol F$ denotes the feature map of the network output, which includes the VIT's class (CLS) token and patch embedding, and $\boldsymbol y$ denotes the quality score.
\subsection{Overview}
In this paper, we present Quality-Aware Feature Matching Image Quality Metric~(QFM-IQM), a novel framework that can effectively distinguish between images with similar semantic features but different quality scores. As depicted in Fig.~\ref{modelStructure}, QFM-IQM seamlessly integrates three main components: {Semantic Noise Matching (SNM)}, {Quality Consistency Constraint (QCC)} and {Distillation Label Extension (DLE)}.
Initially, the input distorted image is fed into a transformer encoder to obtain feature $\boldsymbol F_o \in \mathbb{R}^{(M+1)\times D}$ from the final layer. Subsequently, the SNM module matches these features with those that have similar quality scores but different semantics~(Sec.~\ref{SNM}). Next, the matched semantic noise $\hat{\boldsymbol F}_o$ is introduced as feature perturbation into feature $\boldsymbol F_o$ to obtain feature $\mathcal{F}_o$, and the QCC module constraints between the quality prediction of the $\boldsymbol F_o$ and $\mathcal{F}_o$ to remain consistent, forcing the model to be robust to quality-unrelated semantic noise during adversarial learning~(Sec.~\ref{QCC}). Additionally, the DLE module uses knowledge distillation to generate pseudo-labels for unlabeled samples, thereby enriching the adversarial learning dataset of QCC~(Sec.~\ref{DLE}). These components collectively contribute to a more effective and accurate NR-IQA process. 
It's worth noting that during the inference process, scores are predicted directly using the encoder and decoder, discarding the above three modules, and avoiding any additional computational overhead.
\vspace{-5pt}
\subsection{Architecture Design}\label{backbone}
\noindent \textbf{Transformer Encoder.}
Our model uses a Transformer encoder to process input image patches. These patches are first transformed into D-dimensional embeddings, including a class token and position embeddings for spatial information. Three linear projection layers transform patch embedding into matrices ${\boldsymbol Q, \boldsymbol K, \boldsymbol V} \in \mathbb{R}^{(M+1)\times D}$ for query, key, and value. This operation, combined with Multi-Head Self-Attention~(MHSA), layer normalization, and a Multi-Layer Perceptron (MLP), produces the output feature $\boldsymbol F_o$.

\noindent \textbf{Transformer Decoder.}
Previous works~\cite{DEIQT} found that CLS tokens cannot build an optimal representation for image quality. we utilize a quality-aware decoder that further interprets the CLS token. The decoder uses Multi-Head Cross-Attention (MHCA) with a set of queries $Q_d$, interacting with encoder outputs to calculate the final quality score $\boldsymbol{y}$. This method improves the learning and generalization capabilities of our NR-IQA model.
\vspace{-5pt}
\subsection{Distillation Label Extension}\label{DLE}
Considering the challenges transformer-based models face in learning sufficient downstream task knowledge with insufficient training data, leading to overfitting and performance degradation, this paper proposes a new distillation strategy. This strategy employs an extra unlabeled dataset \(\mathcal{B}\) for feature adversarial learning during training, effectively using information from samples in the expanded dataset with similar quality scores but different features. Specifically, we first use a teacher model generating pseudo labels for each image in \(\mathcal{B}\) and then store in a temporary memory \(\mathcal{B}\). Then, labeled datasets \(\mathcal{A}\) and \(\mathcal{B}\) are input into the student model together, identifying features with similar quality scores but different semantics (Sec.~\ref{SNM}). These features, treated as noise, are added to the image features, enabling the model to learn from both labeled and unlabeled data and improve its generalization ability for images with different quality.
\vspace{-5pt}
\subsection{Semantic Noise Matching}\label{SNM}
The Semantic Noise Matching (SNM) Module, comprising Temporary Memory and Semantic Noise Feature Matching components, functions by aligning input image features with Temporary Memory samples in each training batch, efficiently matching noise samples.

\noindent \textbf{Temporary Memory.}
The objective of the Temporary Memory is to store features and labels of each mini-batch of data, serving as a queryable resource for the SNM module. Specifically, the student feature extractor conducts forward propagation to extract features of distorted images from labeled dataset \(\mathcal{A}\) and stores these features alongside the ground truth of the distorted images in Temporary Memory \(\mathcal{A}\). Concurrently, the student extracts features of distorted images from unlabelled dataset \(\mathcal{B}\), storing them with the pseudo labels of these images in Temporary Memory \(\mathcal{B}\). It is important to note that during the training process, the original memory data is overwritten in each iteration of every mini-batch to reset the Temporary Memory.

\noindent \textbf{Semantic Noise Feature Matching.}
This process involves selecting images with similar quality but different semantics to use as semantic noise features. Then, these noise features are employed as feature perturbations to disrupt semantics while largely preserving quality information. Finally, under the constraint of quality consistency, this effectively reduces sensitivity to semantic noise (referenced in Sec. \ref{QCC}).
Fig. \ref{select} overviews our Semantic Noise Matching (SNM) module. For a mini-batch of distorted input images, Temporary Memory $\mathcal{A}$ and $\mathcal{B}$ store collections of features and labels, then QFM-IQM leverages the SNM module to select noise samples that have similar quality scores but different semantic features from both labeled and unlabeled datasets. Specifically, the process begins by flattening the feature $\boldsymbol{F}_o$ of the input image, obtained from the encoder, into a feature vector $\boldsymbol{f}_o$. Subsequently, another feature vector $\hat{\boldsymbol{f}}_o$ is matched from the Temporary Memory $\mathcal{A}$ or $\mathcal{B}$ based on the matching strategy. The evaluation metric for the matching strategy is formulated as the product of feature similarity and score distance. Feature similarity is calculated using cosine distance, while score distance is determined using the $\mathcal{L}_2$ norm. The mathematical expression is as follows:
\begin{equation}
    \mathcal{S}((\boldsymbol{{f}}_o, \rm{{y}}_o), (\hat{\boldsymbol{f}}_o, \rm{\hat{{y}}_o})) = \frac{\boldsymbol{{f}}_o \cdot \hat{\boldsymbol{f}}_o}{\|\boldsymbol{{f}}_o\| \cdot \|\boldsymbol{\hat{{f}}_o}\|} \cdot \|\rm{{y}}_o - \rm{\hat{{y}}_o}\|.
\end{equation}
Here, $\rm {y}_o$ and $\rm {\hat y}_o$ represent the quality score of the sample, $\cdot$ represents the dot product between the vectors, and $\|\cdot\|$ represents their $\mathcal{L}_2$ norms. This score is then used as a criterion for the final selection strategy, choosing the top K minimum score features selected as matching samples for the following QCC module's input.

\subsection{Quality Consistency Constraint}\label{QCC}
Existing studies have not addressed the challenge of differentiating between images with similar semantic features but different quality scores and eliminating quality-irrelevant features from high-level pre-trained features. To address this, we introduce the Quality Consistency Constraint~(QCC) module to isolate quality-related attributes from an image's semantic content. 
In contrast to previous contrastive learning approaches, this paper introduces a straightforward yet effective adversarial method to enhance a model's robustness to features unrelated to image quality. This is accomplished by strengthening the consistency of the model's predictions for image quality before and after introducing noise to distorted images. This method reduces the model's sensitivity to semantic noise and compels it to exhibit robustness to subtle variations, ultimately increasing the separation between feature representations of noise samples.
Concretely, we first extract the K features with the smallest evaluation metric calculated in the SNM module and incorporate them as feature perturbations into the original features. Since we perform feature selection twice in labeled dataset \(\mathcal{A}\) and unlabelled dataset \(\mathcal{B}\), the expression of this process is:
\begin{equation}\label{contrast}
        {\mathcal{F}_o} = (1-{\lambda _1}-{\lambda _2}){{{\boldsymbol F}}_o} + \frac{\lambda_1}{K}\sum_{j=1}^{K} {\boldsymbol {\hat{F}}^{\mathcal{A}}_{o,j}}  + \frac{\lambda_2}{K}\sum_{j=1}^{K} {\boldsymbol {\hat{F}}^{\mathcal{B}}_{o,j}},
\end{equation}%
where the weight $\lambda _1$ and $\lambda _2$ controls the strength of the perturbation. ${\boldsymbol {\hat{F}}^\mathcal{A}_{o,j}}$ and ${\boldsymbol {\hat{F}}^{\mathcal{B}}_{o,j}}$ represent the $j^{th}$ smallest matching feature from the datasets \(\mathcal{A}\) and datasets \(\mathcal{B}\). 
After that, we feed the feature $\mathcal{F}_o$ which is obtained by the Eq.~\ref{contrast} into the decoder to estimate the quality score $\boldsymbol y$. 
Notably, even after adding noise, the predicted quality is still supervised using the ground truth of the original image. By minimizing the prediction loss on samples with semantic noise, the model's robustness to the semantic noise is improved.
\begin{table*}[t]
\setlength\tabcolsep{0.7pt}
    \centering
      \caption{Performance comparison of average SRCC and PLCC, with bold indicating the best results and \underline{underlines} for the second-best.}  
      
    \resizebox{1\textwidth}{!}{
        \begin{tabular}{lcccccccc||cccccccc}
      \toprule[1.5pt]
    & \multicolumn{2}{c}{LIVE} & \multicolumn{2}{c}{CSIQ} & \multicolumn{2}{c}{TID2013} & \multicolumn{2}{c||}{KADID} & \multicolumn{2}{c}{LIVEC} & \multicolumn{2}{c}{KonIQ} & \multicolumn{2}{c}{LIVEFB} & \multicolumn{2}{c}{SPAQ}\\
    \cmidrule{2-17}    Method & \multicolumn{1}{c}{PLCC} & \multicolumn{1}{c}{SRCC} & \multicolumn{1}{c}{PLCC} & \multicolumn{1}{c}{SRCC} & \multicolumn{1}{c}{PLCC} & \multicolumn{1}{c}{SRCC}& \multicolumn{1}{c}{PLCC} & \multicolumn{1}{c||}{SRCC}& \multicolumn{1}{c}{PLCC} & \multicolumn{1}{c}{SRCC}& \multicolumn{1}{c}{PLCC} & \multicolumn{1}{c}{SRCC}& \multicolumn{1}{c}{PLCC} & \multicolumn{1}{c}{SRCC}& \multicolumn{1}{c}{PLCC} & \multicolumn{1}{c}{SRCC}\\
    \midrule
    DIIVINE ~\cite{saad2012blind} & 0.908 & 0.892 & 0.776 & 0.804 & 0.567 & 0.643 & 0.435 & 0.413 & 0.591 & 0.588 & 0.558 & 0.546 & 0.187 & 0.092 & 0.600 & 0.599 \\
    BRISQUE ~\cite{BRISQUE} & 0.944 & 0.929 & 0.748 & 0.812 & 0.571 & 0.626 & 0.567 & 0.528 & 0.629 & 0.629 & 0.685 & 0.681 & 0.341 & 0.303 & 0.817 & 0.809 \\
    ILNIQE ~\cite{ILNIQE} & 0.906 & 0.902 & 0.865 & 0.822 & 0.648 & 0.521 & 0.558 & 0.534 & 0.508 & 0.508 & 0.537 & 0.523 & 0.332 & 0.294 & 0.712 & 0.713 \\
    BIECON ~\cite{BIECON} & 0.961 & 0.958 & 0.823 & 0.815 & 0.762 & 0.717 & 0.648 & 0.623 & 0.613 & 0.613 & 0.654 & 0.651 & 0.428 & 0.407 & {-} & {-} \\
    MEON ~\cite{MEON} & 0.955 & 0.951 & 0.864 & 0.852 & 0.824 & 0.808 & 0.691 & 0.604 & 0.710  & 0.697 & 0.628 & 0.611 & 0.394 & 0.365 & {-} & {-} \\
    WaDIQaM ~\cite{bosse2017deep} & 0.955 & 0.960  & 0.844 & 0.852 & 0.855 & 0.835 & 0.752 & 0.739 & 0.671 & 0.682 & 0.807 & 0.804 & 0.467 & 0.455 & {-} & {-} \\
    DBCNN ~\cite{zhang2018blind} & 0.971 & 0.968 & {0.959} & {0.946} & 0.865 & 0.816 & 0.856 & 0.851 & 0.869 & 0.851 & 0.884 & 0.875 & 0.551 & 0.545 & 0.915 & 0.911 \\
    MetaIQA ~\cite{zhu2020metaiqa} & 0.959 & 0.960  & 0.908 & 0.899 & 0.868 & 0.856 & 0.775 & 0.762 & 0.802 & 0.835 & 0.887 & 0.850 & 0.507 & 0.54  & {-} & {-} \\
    P2P-BM ~\cite{ying2020patches} & 0.958 & 0.959 & 0.902 & 0.899 & 0.856 & 0.862 & 0.849 & 0.84  & 0.842 & 0.844 & 0.885 & 0.872 & 0.598 & 0.526 & {-} & {-} \\
    HyperIQA ~\cite{hypernet} & 0.966 & 0.962 & 0.942 & 0.923 & 0.858 & 0.840  & 0.845 & 0.852 & 0.882 & 0.859 & 0.917 & 0.906 & 0.602 & 0.544 & 0.915 & 0.911 \\
    TReS  ~\cite{TReS} & 0.968 & 0.969 & 0.942 & 0.922 & 0.883 & 0.863 & 0.858 & 0.859 & 0.877 & 0.846 & {0.928} & 0.915 & 0.625 & 0.554 & {-} & {-} \\
    MUSIQ ~\cite{ke2021musiq} & 0.911 & 0.940  & 0.893 & 0.871 & 0.815 & 0.773 & 0.872 & 0.875 & 0.746 & 0.702 & {0.928} & {0.916} & {0.661} & {0.566} & {0.921} & {0.918} \\
    DACNN ~\cite{pan2022dacnn} & {0.980}  & {0.978} & {0.957} & {0.943} & {0.889} & {0.871} & \underline {0.905} & \underline{0.905} & {0.884} & {0.866} & 0.912 & 0.901 & {-} & {-} & {0.921} & 0.915 \\
    DEIQT ~\cite{DEIQT} & \underline{0.982} & \underline{0.980} & \underline{0.963} & {0.946} & \underline{0.908} & \underline{0.892} & 0.887 & 0.889 & \underline{0.894} & \underline{0.875} & \underline{0.934} & \underline{0.921} & {0.663} & \underline{0.571} & {0.923} & \underline{0.919} \\
    Re-IQA ~\cite{saha2023re} & {0.971}  & {0.970} & {0.960} & \underline{0.947} & {0.861} & {0.804} &  {0.885} & {0.872} & {0.854} & {0.840} & 0.923 & 0.914 & \textbf{0.733} & \textbf{0.645} & \textbf{0.925} & 0.918 \\
    \midrule
    QFM-IQM~(Ours) & \textbf{0.983} & \textbf{0.981} & \textbf{0.965} & \textbf{0.954} & \textbf{0.932} & \textbf{0.916} & \textbf{0.906} & \textbf{0.906} & \textbf{0.913} & \textbf{0.891} & \textbf{0.936} & \textbf{0.922} & \underline{0.667} & {0.567} & \underline{0.924} & \textbf{0.920} \\
    \bottomrule
    \end{tabular}}
    \vspace{-5pt}
  \label{tab:booktabs}
\end{table*}

\begin{table}[t]
\small
\setlength\tabcolsep{3.5pt}
  \centering
    \vspace{-5pt}
    \caption{SRCC on the cross datasets validation. The best performances are highlighted in boldface.}
    
    \begin{tabular}{ccccccc}
    \toprule
    Training & \multicolumn{2}{c}{  LIVEFB } & \multicolumn{1}{c}{LIVEC} & \multicolumn{1}{c}{KonIQ} & \multicolumn{1}{c}{LIVE} & \multicolumn{1}{c}{CSIQ} \\
    \midrule
    Testing & \multicolumn{1}{c}{KonIQ} & \multicolumn{1}{c}{LIVEC} & \multicolumn{1}{c}{KonIQ} & \multicolumn{1}{c}{LIVEC} & \multicolumn{1}{c}{CSIQ} & \multicolumn{1}{c}{LIVE} \\
    \midrule
    DBCNN & 0.716 & 0.724 & 0.754 & 0.755 & 0.758 & 0.877 \\
    P2P-BM &0.755 & 0.738 & 0.74  & 0.77  & 0.712 & {-} \\
    HyperIQA & 0.758 & 0.735 & {0.772} & 0.785 & 0.744 & 0.926 \\
    TReS  & 0.713 & 0.74  & 0.733 & 0.786 & 0.761 & {-} \\
    Re-IQA & - & - & 0.769 & 0.791 & 0.808 & 0.929 \\
    DEIQT & 0.733 & 0.781 & 0.744 & 0.794 & 0.781 & 0.932 \\
    \midrule
    QFM-IQM   & \textbf{0.768} & \textbf{0.791} & \textbf{0.775} & \textbf{0.796} & \textbf{0.820} & \textbf{0.941} \\
    \bottomrule
    \end{tabular}%
  \vspace{-10pt}
  \label{cross}%
\end{table}%

\section{Experiments}
\subsection{Benchmark Datasets and Evaluation Protocols}
We evaluate the QFM-IQM on 8 standard NR-IQA datasets. These include four synthetic datasets: LIVE~\cite{sheikh2006statistical}, CSIQ~\cite{larson2010most}, TID2013~\cite{ponomarenko2015image}, and KADID~\cite{lin2019kadid}; and four authentic datasets: LIVEC~\cite{ghadiyaram2015massive}, KonIQ~\cite{hosu2020koniq}, LIVEFB~\cite{ying2020patches}, and SPAQ~\cite{fang2020perceptual}.
Synthetic datasets are generated by applying distortions like JPEG compression and Gaussian blur to original images. LIVE and CSIQ contain 779 and 866 images with five and six distortion types, respectively, while TID2013 and KADID have 3000 and 10,125 images with 24 and 25 distortion types.
The authentic datasets contain images captured by different photographers using various mobile devices, with LIVEC comprising 1162 images, SPAQ 11,125 images, and KonIQ 10,073 images sourced from public multimedia resources. LIVEFB is the largest authentic dataset with 39,810 images.
The performance of the QFM-IQM model, in terms of prediction accuracy and monotonicity, is assessed using Spearman's Order Correlation Coefficient (SRCC) and Pearson's Linear Correlation Coefficient (PLCC). Both SRCC and PLCC values range from -1 to 1, with values near 1 denoting higher performance in both metrics.
\vspace{-5pt}
\subsection{Implementation Details}
\noindent\textbf{Pre-trained Teacher.} 
Our teacher network, comprising a VIT-S and a decoder, is pre-trained on the largest authentic dataset LIVEFB for offline knowledge distillation on students trained on the IQA dataset. The pre-trained strategy follows the \cite{DEIQT}. 
Notably, when the student model is trained on the LIVEFB dataset, the teacher model is pre-trained on KonIQ to ensure fairness.

\noindent\textbf{Student Training.} 
 To train the student network, we follow the standard approach of cropping input images into ten patches of 224 $\times$ 224 resolution, reshaped into smaller 16 $\times$ 16 patches with a 384-dimensional input token. Using a Transformer encoder based on ViT-S from DeiT III~\cite{touvron2022deit}, our model has 12 layers and 6 heads, coupled with a single-layer decoder. Training is conducted over 9 epochs with a learning rate of $2 \times 10^{-4}$, reducing by a factor of 10 every 3 epochs, using the Adamw optimizer~\cite{loshchilov2017decoupled}. The batch size depends on the size of the dataset, which ranges from 16 to 128. For each dataset, 80\% of the images are used for training and the remaining 20\% of the images are used for testing. This process is repeated ten times to minimize bias. For synthetic distortion datasets, training and testing sets are divided by reference images for content independence. The model's performance is quantified by the average SRCC and PLCC, measuring prediction accuracy and monotonicity.
\vspace{-5pt}
\subsection{Overall Prediction Performance Comparison}
The results of the comparison between QFM-IQM and 15 classical or state-of-the-art~(SOTA) NR-IQA methods, which include hand-crafted feature-based NR-IQA methods like ILNIQE ~\cite{ILNIQE} and BRISQUE ~\cite{BRISQUE}, as well as deep learning-based methods such as MUSIQ ~\cite{ke2021musiq} and DEIQT ~\cite{DEIQT}, are presented in Table \ref{tab:booktabs}. It can be observed on six of the eight datasets that QFM-IQM outperforms all other methods in terms of performance. Achieving competitive performance on all of these datasets is a challenging task due to the wide range of image content and distortion types. Therefore, these observations confirm the effectiveness of QFM-IQM in accurately characterizing image quality.
\subsection{Generalization Capability Validation}
To assess the generalization capabilities of QFM-IQM, we conduct a series of cross-dataset validation experiments. In these tests, QFM-IQM trains on one dataset and then evaluates it on different datasets without adjustments to its parameters. The results, presented in Table \ref{cross}, demonstrate QFM-IQM's superior performance over SOTA models across six datasets. This includes notable gains on the LIVEC dataset and strong results on the KonIQ dataset. These results, benefiting from the noise-resistant training provided by the SNM and QCC, as well as the extensive semantic noise of various distortions provided by the DLE, underscore QFM-IQM's exceptional ability to generalize.
\begin{figure*}[t!]
\centering{\includegraphics[width=1\textwidth]{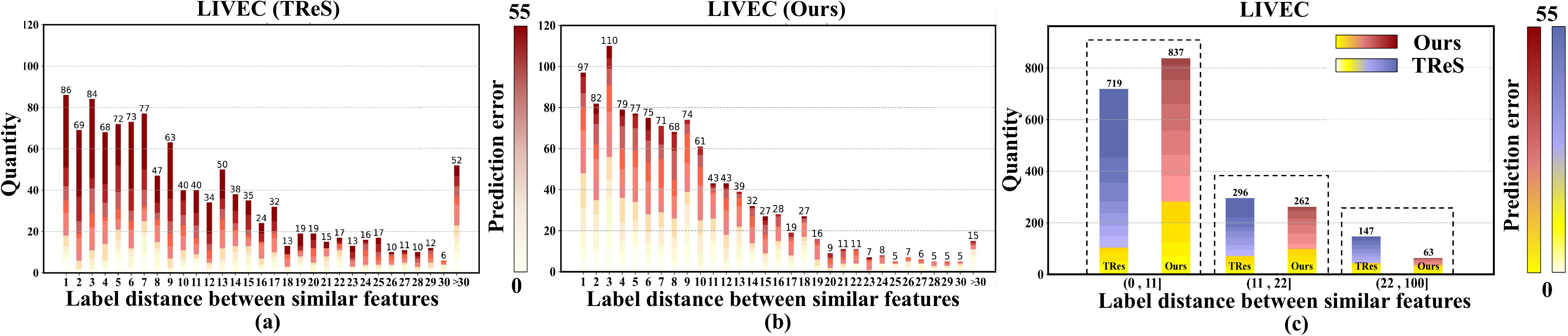}}
 \vspace{-20pt}
	\caption{Qualitative analysis on image pairs have similar semantic features but different quality scores.
 The bar charts in (a) and (b) display the number of image pairs with the \textbf{most similar} features within a specific label distance range. For TReS, there are many image pairs with similar features but large differences in quality scores (even with label distance greater than 30, TReS considers 52 samples as feature-similar). Our model greatly alleviates this issue and achieves a lead in accuracy (a large number of light yellow areas represent smaller prediction errors). In bar char (c), we integrate the statistical results of (a) and (b) to make the above conclusions clearer.
 }
 \label{zhuzhuan}
\end{figure*}

\begin{center} 
\begin{table*}[t]
\caption{Ablation experiments on TID2013, LIVEC and KonIQ datasets. Here, $\mathcal{A}^{QCC}$ and $\mathcal{B}^{QCC}$ refer to the Quality Consistency Constraint without and with the Distillation Label Extension module, where bold entries indicate the best result.}

\setlength\tabcolsep{3.0pt}
    \centering
   \begin{tabular}{ccc|cccccc}
    \toprule 
    \multirow{2}{*}{ Index } & \multirow{2}{*}{$\mathcal{A}^{QCC}$} & \multirow{2}{*}{$\mathcal{B}^{QCC}$} & \multicolumn{2}{c}{TID2013} & \multicolumn{2}{c}{LIVEC}& \multicolumn{2}{c}{KonIQ}
    \\\cmidrule(r){4-5}\cmidrule(r){6-7}\cmidrule(r){8-9}
     & & & \multicolumn{1}{c}{PLCC} & \multicolumn{1}{c}{SRCC}  & \multicolumn{1}{c}{PLCC}  & \multicolumn{1}{c}{SRCC} & \multicolumn{1}{c}{PLCC}  & \multicolumn{1}{c}{SRCC} \\
    \midrule
\textbf{\textit{a)}}       &   &    & 0.899~(±0.017) & 0.891~(±0.028) & 0.883~(±0.012) & 0.865~(±0.021) & 0.926~(±0.003) & 0.913~(±0.003)       \\
\textbf{\textit{b)}}       & \CheckmarkBold  &     & 0.919~(±0.022) & 0.901~(±0.026) & 0.904~(±0.009) & 0.885~(±0.011) & 0.934~(±0.001) & 0.922~(±0.002)       \\
\textbf{\textit{c)}}       &   &  \CheckmarkBold   & 0.925~(±0.016) & 0.910~(±0.022) & 0.907~(±0.006) & 0.882~(±0.010) & 0.935~(±0.002) & 0.921~(±0.002)       \\
\textbf{\textit{d)}}       & \CheckmarkBold  &  \CheckmarkBold   & \textbf{0.932~(±0.015)} & \textbf{0.916~(±0.017)} & \textbf{0.913~(±0.007)} & \textbf{0.891~(±0.009)} & \textbf{0.936~(±0.002)} & \textbf{0.922~(±0.003)}       \\
    \bottomrule
    \end{tabular}
    \label{ablation}
\end{table*}
\end{center}
\vspace{-25pt}
\subsection{Ablation Study}
The core idea of our QFM-IQM is to improve the model's ability to distinguish subtle quality differences by enhancing its robustness to semantic noise. Therefore, we explore the impact of Quality Consistency Constraint~(QCC) operation on the model when compared with different datasets in Table~\ref{ablation}. Let $\mathcal{A}^{QCC}$ and $\mathcal{B}^{QCC}$ denote the QCC operations applied only to the labeled dataset~$\mathcal{A}$ and the unlabeled dataset~$\mathcal{B}$, respectively. In the ablation experiments on the QCC module, we use the Semantic Noise Matching (SNM) module by default.

\noindent \textbf{Effective of QCC on labeled dataset~$\mathcal{A}$.}
We explore the effectiveness of the QCC module on the original training dataset. After using QCC for training, performance on three datasets improves to varying degrees. Particularly on the LIVEC authentic dataset, there is a significant improvement of 2.1 points, and the variance is significantly reduced. However, we find that on the TID2013 synthetic dataset, although there is a significant improvement of 2 points, the stability of the model decreases. We analyze this as due to the synthetic dataset having less rich semantic information than the authentic dataset, causing our QCC module to force the model to ignore the quality-related semantic information, thereby weakening its stability.

\noindent \textbf{Effective of QCC on unlabeled dataset~$\mathcal{B}$.}
We further explore the effectiveness of the QCC module on the expanded unlabeled training dataset. Compared to $\mathcal{A}^{QCC}$, $\mathcal{B}^{QCC}$ achieves more significant performance improvements. Additionally, the issue of decreased stability on the TID2013 dataset is rectified, ensuring stability across all datasets. This indicates that using additional pseudo-labels to expand our dataset allows our QFM-IQM to learn from images with different quality levels and feature distributions. Consequently, our model can distinguish quality-related features through more extensive data comparisons. The combination of $\mathcal{A}^{QCC}$ and $\mathcal{B}^{QCC}$ leads to a superior model for IQA.

\begin{figure*}[htbp]
\centering{\includegraphics[width=171mm,height=72mm]{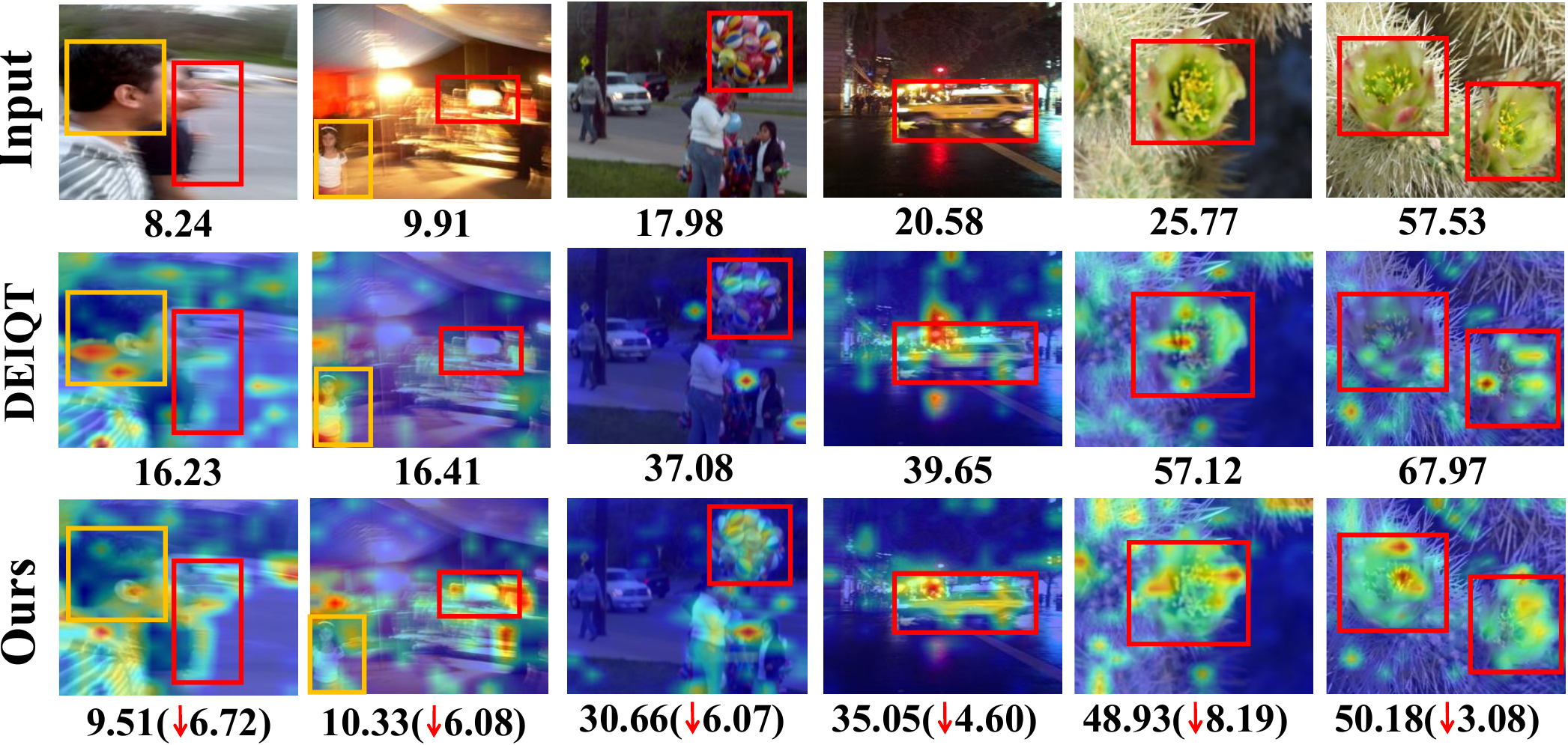}}
 \vspace{-20pt}
	\caption{Activation maps of DEIQT ~\cite{DEIQT} and QFM-IQM using Grad-CAM ~\cite{grad} on authentic dataset LIVEC. The scores below the first row of images represent \emph{Mean Opinion Scores} (MOS). Our model pays more attention to image distortion regions (red boxes) and is less sensitive to semantic noise (yellow boxes), and correspondingly our image quality prediction ability is closer to the true value. Rows 1-3 represent input images, CAMs from DEIQT, and CAMs from QFM-IQM, respectively.
 }
 \label{visualize}
 \vspace{-5pt}
\end{figure*}

\noindent \textbf{Effective on SNM and QCC.}
We extend our study to ablation experiments on SNM and QCC mechanisms, as shown in Table \ref{ablation_sn}. Specifically, we undertake two sets of SNM ablation experiments: the first set, ${\text{SNM}}^F$, matches pairs of samples dissimilar in features, while the second set, ${\text{SNM}}^Q$, matches pairs similar in quality. Additionally, an extra ablation study without using the QCC involves matching noise samples $(\boldsymbol x_1,\boldsymbol y_1)$ and $(\boldsymbol x_2,\boldsymbol y_2)$ to a given pair $(\boldsymbol x,\boldsymbol y)$, where $\boldsymbol y$ is the quality label, adjusting the quality labels by a specific ratio ($\boldsymbol \beta_1 =\boldsymbol \beta_2 = 0.1$) to incorporate noise sample labels, formulated as $\boldsymbol y = (1-\boldsymbol \beta_1-\boldsymbol \beta_2){\boldsymbol y} + {\boldsymbol \beta_1}{\boldsymbol y_1} + {\boldsymbol \beta_2}{\boldsymbol y_2}$.

Our findings underscore the critical role of QCC's constraints because even slight changes in mixing ratios may significantly alter the perceived quality category, potentially misleading the model during the supervision process. Moreover, the SNM mechanism emphasizes the importance of matching samples with similar quality but significantly different features because matching such noise samples can lead to the model making the largest errors in quality prediction. Therefore, considering only one aspect in the matching process may result in suboptimal solutions and fail to effectively improve the model's robustness to noise samples.
\begin{table}[t]
\centering
\setlength\tabcolsep{5.0pt}
    \vspace{-5pt}
\caption{Ablation experiments on LIVEC and TID2013 datasets. Here, ${\text{SNM}}^Q$ and ${\text{SNM}}^F$ refer to the SNM module only matching pairs with similar quality and dissimilar features, respectively.}
\renewcommand{\arraystretch}{0.4}
\resizebox{0.48\textwidth}{!}{
\begin{tabular}{ccc|cccc} 
\toprule 
\multicolumn{3}{c|}{Component} & \multicolumn{2}{c}{LIVEC} & \multicolumn{2}{c}{TID2013}  \\ 
\midrule
${\text{SNM}}^Q$ & ${\text{SNM}}^F$ & QCC          & PLCC  & SRCC              & ~PLCC & SRCC                 \\ 
\midrule
       &        &              & 0.883 & 0.865             & 0.899 & 0.891                \\
\CheckmarkBold      &        &              & 0.897 & 0.880             & 0.918 & 0.899                \\
       & \CheckmarkBold      &              & 0.898 & 0.879             & 0.916 & 0.897                \\
\CheckmarkBold      & \CheckmarkBold      &              & 0.901 & 0.882             & 0.921 & 0.907                \\
\CheckmarkBold      &        & \CheckmarkBold            & 0.906 & 0.885             & 0.922 & 0.902                \\
       & \CheckmarkBold      & \CheckmarkBold            & 0.904 & 0.884             & 0.920 & 0.900                \\ 
\midrule
\CheckmarkBold      & \CheckmarkBold      & \CheckmarkBold            & \textbf{0.913} & \textbf{0.891}             & \textbf{0.932} & \textbf{0.916}                \\
\bottomrule
\end{tabular}}
\vspace{-10pt}
\label{ablation_sn}
\end{table}
\begin{table}[t]
\centering
\caption{Ablation experiments on the LIVEC and TID2013 datasets for unlabeled data selection in the DLE module.}
\resizebox{0.48\textwidth}{!}{
\begin{tabular}{c|cccc} 
\toprule
\multirow{2}{*}{Unlabeled Dataset} & \multicolumn{2}{c}{LIVEC} & \multicolumn{2}{c}{TID2013}  \\ 
\cmidrule{2-5}
                                   & PLCC  & SRCC              & PLCC  & SRCC                 \\ 
\midrule
-                                  & 0.904 & 0.885             & 0.919 & 0.901                \\
KADID                              & 0.904 & 0.887             & 0.921 & 0.899                \\
KonIQ                              & 0.912 & 0.889             & 0.928 & 0.913                \\
LIVEFB                             & \textbf{0.913} & \textbf{0.891}             & \textbf{0.932} & \textbf{0.916}                \\
\bottomrule
\end{tabular}}
\label{ablation_unlabeled}
\end{table}

\vspace{-5pt}
\noindent \textbf{Ablation study of unlabeled dataset selection.}
We explore using the synthetic dataset KADID (10,125 images) and the authentic datasets KonIQ (10,073 images) and LIVEFB (39,810 images) as unlabeled data sources, as summarized in Table \ref{ablation_unlabeled}. The results indicate more significant improvements with richer unlabeled datasets (e.g., on the TID2013 dataset, LIVEFB improves by 1.3\%, whereas KADID and KonIQ improve by 0.2\% and 0.9\%, respectively). We draw two conclusions:
(1) Increasing the volume of unlabeled data enhances performance. As data becomes more diverse, unlabeled datasets provide a wider variety of adversarial noise samples, compelling the model to improve its robustness against a broader range of noise features.
(2) Authentic datasets offer more significant benefits than synthetic datasets. Specifically, synthetic datasets are often generated by applying various degrees of distortion to a single original image. For instance, the KADID dataset, with 10,125 distorted images, contains only 81 unique content images, implying that synthetic datasets offer limited semantic information. Consequently, models struggle to effectively learn robustness against semantic noise from synthetic datasets.
\begin{table}[t]
\caption{Analysis of the K in memory searching on both authentic and synthetic datasets. Bold entries indicate the best performance.}
\setlength\tabcolsep{10pt}
\renewcommand{\arraystretch}{0.85} 
\centering
\begin{tabular}{ccccc}
\toprule
& \multicolumn{2}{c}{LIVE} & \multicolumn{2}{c}{LIVEC} \\
\cmidrule{2-5}
K-NN & \multicolumn{1}{c}{PLCC} & \multicolumn{1}{c}{SRCC} & \multicolumn{1}{c}{PLCC} & \multicolumn{1}{c}{SRCC} \\
\midrule
{K=1} & \textbf{0.983} & \textbf{0.981} & 0.910 & 0.887 \\
{K=2} & 0.981 & 0.978 & 0.905 & 0.887 \\
{K=3} & 0.982 & 0.979 & 0.900 & 0.880 \\
{K=4} & 0.982 & 0.980 & 0.904 & 0.885 \\
{K=6} & 0.981 & 0.978 & \textbf{0.913} & \textbf{0.891} \\
{K=8} & 0.978 & 0.976 & 0.903 & 0.877 \\
\bottomrule
\end{tabular}%
  \vspace{-10pt}
\label{table4}
\end{table}
\subsection{Qualitative Analysis}
\noindent \textbf{Analyze similar features that have different quality.}\label{challenge}
In this section, we illustrate our motivation by counting the number of most similar features extracted by the TReS and QFM-IQM models from image pairs with different quality score differences.
As shown in Fig. \ref{zhuzhuan}, the vertical axis represents quantity, and the horizontal axis represents the range of label differences. According to Fig. \ref{zhuzhuan}(a), we observe that SOTA pre-trained models like TReS struggle to distinguish feature distances between image pairs with different quality scores.
Specifically, in Fig. \ref{zhuzhuan}(a), TReS incorrectly identifies 52 image pairs with a quality score difference greater than 30 as having very similar features. In this paper, We believe this is due to feature confusion, where the model is misled by similar semantics, causing it to perceive similar features in images with different quality scores. This leads to incorrect quality predictions. Our approach mitigates this issue by reducing the model's sensitivity to semantic noise. Fig. \ref{zhuzhuan}(c) summarizes (a) and (b), showing our model's effectiveness. Compared to TReS, the QFM-IQM model reduces the number of image pairs with similar features but different scores by approximately 56\%, while increasing pairs with close scores and similar features by about 16\%. This indicates a better positive correlation between features and quality scores, enhancing overall prediction performance. Additionally, QFM-IQM shows fewer high prediction errors (fewer instances in red), further proving its effectiveness. In conclusion, our model emphasizes quality-related features more than TReS, enhancing quality-aware features and significantly reducing quality estimation bias.

\noindent \textbf{Visualization of attention map.}
We utilize GradCAM \cite{grad} to visualize the feature attention maps in Fig. \ref{visualize}. The results show that our model accurately and comprehensively focuses on the distorted regions (including both quality-related semantic degradations and low-level artifacts), while DEIQT \cite{DEIQT} fails to accurately pinpoint the true areas of quality degradation. This indicates that our model excels in learning the subtle differences between semantic structures and quality-aware features. We achieve this by filtering out features irrelevant to quality from the pre-trained model, focusing on learning those quality-aware features. It is worth noting that for overall blurry images, such as those in the third column, our model tends to focus on subject areas that are inherently more prone to becoming blurry, such as floating balloons.
\begin{table}[t]
  \caption{Sensitivity study of hyperparameters ${\lambda _1}$ on LIVE, LIVEC, and KonIQ datasets. Bold entries indicate the best performance.} 
\setlength\tabcolsep{4.5pt}
  \centering
    \begin{tabular}{lcccccc}
    \toprule
    & \multicolumn{2}{c}{LIVE} & \multicolumn{2}{c}{LIVEC} & \multicolumn{2}{c}{KonIQ} \\
    \cmidrule{2-7}    ${\lambda _1}$ & \multicolumn{1}{c}{PLCC} & \multicolumn{1}{c}{SRCC} & \multicolumn{1}{c}{PLCC} & \multicolumn{1}{c}{SRCC} & \multicolumn{1}{c}{PLCC} & \multicolumn{1}{c}{SRCC} \\
    \midrule
    E-1   & 0.978 & 0.976& 0.884 & 0.857 & 0.925 & 0.911 \\
    E-2    &0.979 &0.976  & 0.877 & 0.865 & 0.934  & 0.919 \\
    E-4    & 0.979 & 0.979 & 0.897 & 0.875 & 0.931  & 0.916 \\
    E-5    & 0.980 & 0.978 & 0.904 & 0.884 & 0.932  & 0.917 \\
    E-7    & \textbf{0.983} & \textbf{0.981} & \textbf{0.910} & 
    \textbf{0.887} & \textbf{0.936} & \textbf{0.922} \\
    E-8    & 0.981 & 0.979 & 0.902 & 0.886 & 0.933  & 0.917 \\
    \bottomrule
    \end{tabular}%
    \vspace{-11pt}
  \label{hyperparameters_1}
\end{table}%

\vspace{-5pt}
\subsection{Analysis of the K in Memory Searching}
\vspace{-5pt}
In this analysis, we examine the effect of the memory searching operation's searching number, K. We conduct experiments on both synthetic and authentic datasets, LIVE and LIVEC, with K values of 1, 2, 3, 4, 6, and 8. The results, as shown in Table \ref{table4}, demonstrate that the performance on holistic IQA datasets remains stable with various K values, with only a slight fluctuation within 1\%.
Specifically, in the synthetic dataset LIVE, which comprises only 30 distinct content images, the types of distortions (like Gaussian blur and JPEG) are generally uniform across the whole image. Thus, the model receives relatively less contamination at the semantic level, and setting K to 1 achieves the best results. On the other hand, in the authentic dataset LIVEC, all the training data contain different contents, leading to various semantic level degradations. Therefore, effectively utilizing quality perception features from pre-trained semantic features is particularly important, and we find that the best performance is achieved by setting the K value to 6. However, setting the K value too high can be detrimental. Excessive noise can distort the original features of the image that the model has learned.
\begin{table}[t]
  \caption{Sensitivity study of hyperparameters ${\lambda _2}$ on LIVEC and TID2013 datasets. Bold entries indicate the best performance.} 
    \setlength\tabcolsep{10pt}
    \renewcommand{\arraystretch}{0.9} 
    \centering
  \resizebox{0.48\textwidth}{!}{
    \begin{tabular}{lcccccc}
\toprule
         & \multicolumn{2}{c}{LIVEC} & \multicolumn{2}{c}{TID2013}  \\ 
\cmidrule{2-5}
${\lambda _2}$       & PLCC  & SRCC              & PLCC  & SRCC                 \\ 
\midrule
E-1      & 0.909 & 0.890              & 0.914 & 0.893                \\
E-2      & 0.902 & 0.884             & 0.918 & 0.906                \\
E-4      & 0.911 & \textbf{0.893}             & 0.927 & 0.911                \\
E-5      & 0.912 & \textbf{0.893}             & \textbf{0.932} & 0.914                \\
E-7      & \textbf{0.913} & {0.891}             & \textbf{0.932} & \textbf{0.916}                \\
E-8      & 0.905 & 0.885             & 0.918 & 0.900                  \\
\bottomrule
\end{tabular}}
\vspace{-10pt}
\label{hyperparameters_2}
\end{table}
\vspace{-3pt}
\subsection{Sensitivity study of hyperparameters}
\vspace{-5pt}
In this paper, we use ${\lambda _1}$ and ${\lambda _2}$ in Eq.~\ref{contrast} to balance the contrastive learning and distillation, respectively. To this end, we conduct a sensitivity study on different feature perturbation weights ${\lambda_1}$ and ${\lambda_2}$ to explore the effect of QCC, maintaining the search number K at 1.
As shown in Table \ref{hyperparameters_1} and Table \ref{hyperparameters_2}, our findings reveal that the QCC module has relative sensitivity to the hyperparameters. Specifically, small values of $\lambda$ weaken our QCC's impact, whereas large values of $\lambda$ cause significant changes in the feature space and result in performance degradation.

 \vspace{-5pt}
\section{Conclusion}
 \vspace{-5pt}
In this paper, we introduce QFM-IQM, which tackles the challenge of differentiating images with similar semantics but varying quality scores. 
This is accomplished through the SNM module, which selects images with similar quality scores but notable feature differences as perturbations. The QCC module reduces the model's sensitivity to these perturbations, enabling the model to isolate quality-related attributes from an image's semantic content. Additionally, QFM-IQM uses distillation learning to enhance the dataset with authentic images, improving the model's generalization across different scenarios. Experiments on eight benchmark IQA datasets demonstrate the effectiveness of this approach.

\vspace{-5pt}
\section*{Acknowledgements}
This work was supported by National Science and Technology Major Project (No. 2022ZD0118202), the National Science Fund for Distinguished Young Scholars (No.62025603), the National Natural Science Foundation of China (No. U21B2037, No. U22B2051, No. 62176222, No. 62176223, No. 62176226, No. 62072386, No. 62072387, No. 62072389, No. 62002305 and No. 62272401), and the Natural Science Foundation of Fujian Province of China (No.2021J01002,  No.2022J06001).

\vspace{-5pt}
\section*{Impact Statement}
This paper presents work whose goal is to advance the field of Image quality assessment.  There are many potential societal consequences of our work, none of which we feel must be specifically highlighted here.

\bibliography{main}
\bibliographystyle{icml2024}

\newpage
\appendix
\twocolumn
\section{Appendix Overview}
The supplementary material is organized as follows:
\textbf{Qualitative Analysis} provides more analysis of the effectiveness of the QFM-IQM and is further illustrated in conjunction with our motivation.
\textbf{Training and Evaluation Details:} shows more training and evaluation details. 
\textbf{Comparison with CONTRIQUE}: provides more analysis of existing methods CONTRIQUE~\cite{contrastive} based on contrastive learning.

\section{Qualitative Analysis}
In this section, we provide an additional explanation about the content of Fig.~\ref{zhuzhuan} to help readers understand our motivation.  Our motivation stems from a phenomenon we discovered through experimentation: existing state-of-the-art (SOTA) models based on pre-training, such as TReS, struggle to adequately distinguish the feature distances between image pairs with different quality scores.  This results in inaccurate quality predictions.  As shown in Fig.~\ref{zhuzhuan}(a), TReS mistakenly identifies image pairs with a quality score difference greater than 30 as having very similar features.  We believe this may be due to feature confusion, where current models are misled by the similar semantics of the image pairs, leading them to perceive similar features in images with different quality scores, thereby failing to adequately separate their features.  This results in erroneous quality predictions.  Our method addresses this feature confusion by specifically designing for it.  We select noisy samples with different semantic features but similar quality scores, and through noise perturbation, force the model to reduce its sensitivity to semantic noise.  This effectively increases the feature distance between image pairs with similar semantics but different quality scores.  As illustrated in Fig.~\ref{zhuzhuan}(b), our QFM-IQM significantly mitigates the aforementioned issue.  Specifically, it alleviates the situation where image pairs with a quality score difference greater than 30 still share very similar features, and the correlation between features and quality scores is more positively aligned, meaning similar quality scores have similar quality features.  Additionally, the error in our quality predictions is smaller, further demonstrating the effectiveness of our approach.

\section{Training and Evaluation Details}\label{B}
\textbf{Efficacy of subcomponents in the training phase:} Our \textbf{Distilled Label Expansion} augments the dataset using authentic images of different quality levels on LIVEFB and KonIQ to overcome the lack of sufficient training data in IQA. 
Our \textbf{Semantic Noise Feature Matching} compares the features of input with other features to match similar quality features but different semantics~(Fig.3 of the main paper). These matched features are considered noises to be added to the original features in the process of \textbf{Quality Consistency Constraint}.
Our Quality Consistency Constraint constrains the model to still be supervised by the ground truth for the quality prediction of the noise-added samples. This enables the model to emphasize the subtle differences among distortion images, while concurrently reducing the significance placed on semantic aspects.

\noindent\textbf{Unlabeled data:} Our unlabeled data comes from the LIVEFB and KonIQ datasets, but we do not use the label information provided by these datasets during training. During training, we randomly sample a mini-batch of unlabeled images for training.
\begin{table}[htbp]
\setlength\tabcolsep{0.6pt}
 \caption{Training preprocessing details of selected BIQA datasets.}
\vskip 0.15in
  \centering
  \resizebox{0.47\textwidth}{!}{
    \begin{tabular}{lcccc}
    \toprule
    Dataset & \multicolumn{1}{c}{Resolution} & \multicolumn{1}{c}{Resize}& \multicolumn{1}{c}{Batch Size} & \multicolumn{1}{c}{Label Range}\\
    \midrule
    LIVE & $768\times512$ & $512\times384$ &12 & DMOS [0,100] \\
    CSIQ & $512\times512$  & $512\times512$ &12 & DMOS [0,1] \\
    TID2013 & $512\times384$ &$512\times384$ & 48 & MOS [0,9] \\
    KADID  & $512\times384$ &$512\times384$ & 128  & MOS [1,5] \\
    \midrule
    LIVEC & $500P\sim640P$ & $500P\sim640P$ & 16 & MOS [1,100] \\
    KonIQ   & $768P$ & $512\times384$ & 128 & MOS [0,5] \\
    LIVEFB  & $160P\sim700P$ & $512\times512$ &128 & MOS [0,100] \\
    SPAQ  & $1080P\sim4368P$ & $512\times384$ & 128 & MOS [0,100] \\
    \bottomrule
    \end{tabular}
    }%
  \label{tab1}%
\end{table}%
\noindent\textbf{Implement details of training phrase:} To train the student network, we adopt a standard approach of randomly cropping input images into ten patches, each with a $224 \times 224$ resolution. Subsequently, we reshape these patches into a sequence of smaller patches with a patch size of p = 16 and an input token dimension of D = 384. Furthermore, we present additional training preprocessing details for various datasets, which are not included in the main paper, in Table~\ref{tab1}. For different benchmarks, we employ different training settings for a fair comparison.
\begin{table}[htbp]
\setlength\tabcolsep{10.7pt}
  \caption{Performance comparison measured by averages of SRCC and PLCC, where bold entries indicate the best results.} 
  \vskip 0.15in
    \centering
    \resizebox{0.47\textwidth}{!}{
        \begin{tabular}{lcccc}
      \toprule
    & \multicolumn{2}{c}{KonIQ} & \multicolumn{2}{c}{LIVEFB} \\
    \cmidrule(lr){2-3} \cmidrule(lr){4-5}
    Method & \multicolumn{1}{c}{PLCC} & \multicolumn{1}{c}{SRCC} & \multicolumn{1}{c}{PLCC} & \multicolumn{1}{c}{SRCC} \\
    \midrule
    QFM-IQM & \textbf{0.936} & \textbf{0.922} & \textbf{0.667} & \textbf{0.567} \\
    Teacher & 0.930 & 0.914 & \textbf{0.667} & {0.566} \\
    \bottomrule
    \end{tabular}}
  \label{teacher performance}
\end{table}
\noindent\textbf{Teacher performance:} 
We ensured the high accuracy of the pre-trained teacher model on the dataset to provide reliable pseudo-labels, as shown in Table \ref{teacher performance}.

\noindent\textbf{Inference phrase:} 
During the inference process, we only employ a trained encoder-decoder architecture to directly assess the image's quality which benefits from the QFM-IQM to precisely isolate quality-aware features within the pre-trained semantic feature space, thereby enhancing the accuracy of our quality predictions.

\begin{table*}[htbp]
\setlength\tabcolsep{0.7pt}
  \caption{Performance comparison measured by averages of SRCC and PLCC, where bold entries indicate the best results.} 
  \vskip 0.15in
    \centering
    \resizebox{1\textwidth}{!}{
        \begin{tabular}{lcccccccc||cccccccc}
      \toprule[1.5pt]
    & \multicolumn{2}{c}{LIVE} & \multicolumn{2}{c}{CSIQ} & \multicolumn{2}{c}{TID2013} & \multicolumn{2}{c||}{KADID} & \multicolumn{2}{c}{LIVEC} & \multicolumn{2}{c}{KonIQ} & \multicolumn{2}{c}{LIVEFB} & \multicolumn{2}{c}{SPAQ}\\
    \cmidrule{2-17}    Method & \multicolumn{1}{c}{PLCC} & \multicolumn{1}{c}{SRCC} & \multicolumn{1}{c}{PLCC} & \multicolumn{1}{c}{SRCC} & \multicolumn{1}{c}{PLCC} & \multicolumn{1}{c}{SRCC}& \multicolumn{1}{c}{PLCC} & \multicolumn{1}{c||}{SRCC}& \multicolumn{1}{c}{PLCC} & \multicolumn{1}{c}{SRCC}& \multicolumn{1}{c}{PLCC} & \multicolumn{1}{c}{SRCC}& \multicolumn{1}{c}{PLCC} & \multicolumn{1}{c}{SRCC}& \multicolumn{1}{c}{PLCC} & \multicolumn{1}{c}{SRCC}\\
    \midrule
    CONTRIQUE & 0.961 & 0.960 & 0.955 & 0.942 & 0.857 & 0.843 & \textbf{0.937} & \textbf{0.934} & 0.857 & 0.845 & 0.906 & 0.894 & 0.641 & 0.580 & {0.919} & {0.914} \\
    \midrule
    QFM-IQM(ours) & \textbf{0.983} & \textbf{0.981} & \textbf{0.965} & \textbf{0.954} & \textbf{0.932} & \textbf{0.916} & {0.906} & {0.906} & \textbf{0.913} & \textbf{0.891} & \textbf{0.936} & \textbf{0.922} & \textbf{0.667} & \textbf{0.567} & \textbf{0.924} & \textbf{0.920} \\
    \bottomrule
    \end{tabular}}
  \label{performance}
\end{table*}

\begin{table}[t]
\setlength\tabcolsep{10.8pt}
    \centering
      \caption{SRCC on the cross datasets validation. The best results are highlighted in bold.}
      \vskip 0.15in
    \resizebox{0.47\textwidth}{!}{
    \begin{tabular}{lcccc}
    \toprule
    Training & Testing  & CONTRIQUE & QFM-IQM \\\midrule
    LIVEC & KonIQ & 0.676 & \textbf{0.775}\\
    KonIQ & LIVEC & 0.731 & \textbf{0.796} \\
    LIVE & CSIQ & \textbf{0.823} & 0.820\\ 
    CSIQ & LIVE & 0.925 & \textbf{0.941}\\ \bottomrule
   \end{tabular} }
  \label{cross_con}
\end{table}%

\section{Comparison with CONTRIQUE}
In CONTRIQUE, the approach revolves around treating the IQA representation learning problem as a classification task. In this framework, images are categorized into various classes based on distortion types, and contrastive loss is utilized to learn distinctive features that can discriminate between these classes. However, a limitation of this approach is that each real image is treated as a separate class, which limits the effective mining of information from distorted images with similar quality.
In contrast, our QFM-IQM is not confined to a specific type of distortion. Instead, we approach learning quality-aware features from a different perspective, aiming to reduce the sensitivity of the pre-trained network to semantic content through adversarial learning. This technique effectively separates the attributes related to image quality from the semantic features.
As shown in Table \ref{cross_con} and Table \ref{performance}, QFM-IQM exhibits excellent performance in both comparative and cross-dataset validation experiments.


\end{document}